\documentclass[letterpaper, 10 pt, conference]{ieeeconf}

\IEEEoverridecommandlockouts                              

\usepackage{graphicx}
\usepackage{amsmath}
\usepackage{amssymb}
\usepackage{multirow}
\usepackage{booktabs}
\usepackage{subcaption}
\usepackage{algorithm}
\usepackage{algorithmic}
\usepackage{url}
\usepackage{physics}

\usepackage{censor}

\StopCensoring

\title{\LARGE \bf
Temporal Policy: History-Initialized Action Generation for Robotic Learning from Demonstration
}

\author{\censor{Dylan Miller}$^{1}$ and \censor{Martin Jagersand}$^{1}$
   \thanks{$^{1}$\censor{Dylan Miller} and \censor{Martin Jagersand} are with \censor{Department of Computing Science},
\censor{University of Alberta, Canada} \censor{{\tt\small \{djm2; mj7\}@ualberta.ca}}}%
}

\begin{document}

\maketitle
\thispagestyle{empty}
\pagestyle{empty}

\begin{abstract}
   Generative models excel at capturing multimodal behaviors for robotic Learning from Demonstration (LfD), but often suffer from high inference cost.
   By relying on independent couplings from uninformative Gaussian priors, standard diffusion and flow matching models are forced to learn complex, high-cost vector fields to reach the physical action space.
   This paper introduces Temporal Policy, a generative framework based on stochastic interpolants that formulates action generation as a temporally coupled transport problem.
   By initializing the generative flow at the robot's recent history, we explicitly couple past states to future action sequences.
   This data-dependent coupling reduces transport cost and produces straight vector fields.
   We validate Temporal Policy across visuomotor simulation benchmarks and on a physical Barrett WAM 2x 7DoF teleoperation platform.
   Our approach reduces transport costs by nearly an order of magnitude compared to noise-initialized baselines, achieving a 19.1 ms inference latency on a single NVIDIA RTX 4080.
   Crucially, these geometric and computational efficiencies are achieved while matching the success rates of state-of-the-art baselines.
   This simplified transport geometry bypasses the computational bottleneck of independent Gaussian priors, helping enable high-frequency, closed-loop control.
   The code is publicly available at \url{https://github.com/dmiller12/TemporalPolicy}.
\end{abstract}

\section{INTRODUCTION}

\begin{figure*}[t]
   \centering
   \includegraphics[width=\textwidth]{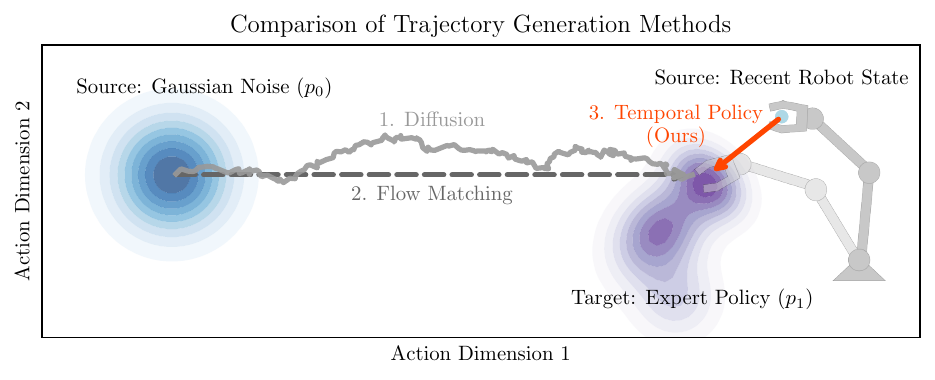}
   \caption{Conceptual comparison of generative modeling approaches.
      (1) \textbf{Diffusion Models} generate stochastic, curved paths from Gaussian noise ($p_0$).
      (2) \textbf{Standard Flow Matching} uses straight paths but still initializes from independent noise.
      (3) Our \textbf{Temporal Policy} initializes the generative process directly to the robot's history, reducing the transport distance.}
   \label{fig:trajectory_comparison}
\end{figure*}

Learning from Demonstration (LfD) allows robots to acquire behaviors directly from human supervision without the need for manual programming or reward engineering \cite{argallSurveyRobotLearning2009}.
These demonstrations can take various forms, ranging from third-person video observation \cite{jinGeneralizableTaskRepresentation2022a} to bilateral teleoperation.
To translate these demonstrations into robust control policies, recent formulations utilize generative models, specifically diffusion \cite{chiDiffusionPolicyVisuomotor2025} and flow matching \cite{lipmanFlowMatchingGenerative2023b}, to capture the rich, multimodal action distributions inherent in human behavior.
However, the standard formulation of these models typically generates data by initializing samples from an uninformative Gaussian noise distribution \cite{hoDenoisingDiffusionProbabilistic2020}.
This approach forces the network to bridge a significant geometric gap between a high-entropy prior and the physical configuration space, resulting in long and potentially high-curvature transport paths.
This transport complexity acts as a fundamental bottleneck, imposing computational burdens and inference latency that limit deployment on real-time, resource-constrained robotic systems.

This paper introduces Temporal Policy, an action generation framework built on the theory of stochastic interpolants \cite{albergoStochasticInterpolantsUnifying2025}.
Instead of initializing from standard Gaussian noise, Temporal Policy formulates action generation as a Point-to-Distribution transport problem initialized directly at the robot's recent observation history (Figure \ref{fig:trajectory_comparison}).
To enable this mapping, the formulation requires state and actions to share the same representation (e.g., joint positions or Cartesian poses).
This fundamentally simplifies the learning objective by reducing the transport cost and exploiting the strong temporal correlation between consecutive states.
Furthermore, analytic score recovery \cite{chenProbabilisticForecastingStochastic2024} enables a single trained model to adjust its noise schedule at inference time, supporting both low-latency deterministic sampling and robust stochastic sampling without retraining.

The primary contributions of this work are:
\begin{itemize}
   \item \textbf{Temporal Coupling Formulation:}
         A data-dependent generative modeling policy that explicitly couples past state history to future actions, reducing transport costs and vector field curvature.
   \item \textbf{Computational Efficiency:}
         By simplifying the learned vector field, the framework requires fewer function evaluations and achieves a $19.1$~ms inference latency, enabling high-frequency real-time control.
   \item \textbf{Empirical Validation:}
         Evaluation across the Robomimic simulation benchmarks, demonstrating competitive performance and superior data efficiency, alongside successful real-world deployment on a physical 7-DoF teleoperation platform.
\end{itemize}

\section{RELATED WORK}

\subsection{Diffusion Based Approaches}
Diffusion based approaches have become popular in LfD largely due to their ability to capture multimodal action distributions.
Diffusion Policy \cite{chiDiffusionPolicyVisuomotor2025} established the efficacy of Denoising Diffusion Probabilistic Models (DDPMs) for visuomotor control, generating temporally coherent action chunks.
Subsequent work has extended this paradigm to continuous-time formulations, leveraging classifier-free guidance to enable goal-conditioned policies \cite{reussGoalConditionedImitationLearning2023}.
However, standard diffusion defines its forward process as a perturbation to an uninformative Gaussian prior \cite{hoDenoisingDiffusionProbabilistic2020}.
Reversing this process requires the network to map from high-entropy noise to highly structured configuration spaces, resulting in long, high-curvature transport paths.
This geometric complexity requires a large number of numerical integration steps and high capacity networks, imposing inference latency that bottlenecks high-frequency closed-loop control.

\subsection{Flow Matching and Optimal Transport}

To improve sampling efficiency, frameworks such as Conditional Flow Matching (CFM) \cite{lipmanFlowMatchingGenerative2023b} and Rectified Flows \cite{liuFlowStraightFast2022} formulate generation as a deterministic Ordinary Differential Equation (ODE).
By constructing linear conditional probability paths, these methods provide a simulation-free regression objective that reduces the required number of numerical integration steps compared to iterative denoising.
Similarly, Consistency Models \cite{songConsistencyModels2023} leverage these principles by learning to map points along the ODE trajectory directly to the data distribution, enabling rapid, few-step generation.

These principles have been adapted for robotic manipulation \cite{funkActionFlowEquivariantAccurate2024, black$p_0$VisionLanguageActionFlow2024}.
Notably, Flow Policy \cite{fangFlowPolicyGeneralizable2026} demonstrates that flow matching can surpass standard diffusion models in both success rate and parameter efficiency by employing specialized transformer architectures (UDiT) and higher-order trigonometric noise schedulers.
Concurrently, frameworks such as ManiFlow \cite{yanManiFlowGeneralRobot2025} have adapted consistency distillation to robotic control to further accelerate inference.

The efficiency of these flows depends heavily on the coupling geometry between the source and target distributions.
To further straighten the learned vector field, techniques such as Optimal Transport CFM (OT-CFM) construct minibatch couplings to approximate the 2-Wasserstein distance, which minimizes path crossings \cite{pooladianMultisampleFlowMatching2023, tongImprovingGeneralizingFlowbased2024}.
Recent work has leveraged these minibatch OT formulations to accelerate training and inference specifically for continuous manipulation tasks \cite{sochopoulosFastFlowbasedVisuomotor2025}.
In stochastic regimes, this optimal transport problem is generalized by the Schr\"{o}dinger Bridge (SB), which recovers the most likely dynamic evolution between two boundary distributions subject to a reference prior \cite{debortoliDiffusionSchrodingerBridge2021, chenLikelihoodTrainingSchrodinger2022}.
While recent advancements have enabled simulation-free SB training for high-dimensional generative modeling \cite{tongSimulationFreeSchrodingerBridges2024}, adapting these frameworks for robotic control remains computationally prohibitive.
Deploying SBs typically requires iterative optimization or complex dual-network architectures to simulate the forward and backward diffusion processes, leading to training instability.

Furthermore, despite these optimizations, the aforementioned flow matching frameworks and their robotic applications fundamentally rely on an independent Gaussian noise source.
In robotic control, where the agent's initial physical state is explicitly known, generating trajectories from uninformative noise ignores critical contextual priors and forces the network to capture an unnecessarily long-distance transport mapping.

\subsection{Data-Dependent Couplings}
While standard generative policies rely on independent Gaussian priors, the CFM framework allows for the construction of vector fields between arbitrary source and target distributions \cite{lipmanFlowMatchingGenerative2023b}.
The Stochastic Interpolants framework provides a further generalization of this approach, unifying diffusion and flow matching by proving that generative Stochastic Differential Equations (SDEs) and ODEs can be derived for any choice of source, target, and interpolant \cite{albergoStochasticInterpolantsUnifying2025}.

This flexibility enables data-dependent initializations, replacing noise with physics-informed priors or degraded states \cite{albergoStochasticInterpolantsDataDependent2024, brinkeSTFlowDataCoupledFlow2026}.
For instance, in general sequence modeling, Trajectory Flow Matching conditions generation on observation history, however, it relies on a dual-network architecture to explicitly parameterize the diffusion process \cite{zhangTrajectoryFlowMatching2024}.
When adapting these data-dependent principles to robotics, recent methods introduce domain-specific compromises.
Streaming Flow Policy initializes flow from the previous action to reduce latency, but targets per-timestep marginals rather than joint distribution over the action sequence, which can cause the policy to artificially mix distinct behaviors during execution \cite{jiangStreamingFlowPolicy2025}.
VITA \cite{gaoVITAVisiontoActionFlow2026} proposed using latent visual representations as the generative source, however they required backpropagating losses through the ODE solver to learn an effective decoder.

We propose Temporal Policy to bypass these cumulative limitations, leveraging the framework proposed by Chen et al.~\cite{chenProbabilisticForecastingStochastic2024}.
This approach maps a point mass centered at the current observation onto a probabilistic ensemble of forecasts via artificial stochastic dynamics.
Such point-to-distribution transport is advantageous because the drift coefficient is non-singular and can be learned efficiently via simulation-free square loss regression \cite{chenProbabilisticForecastingStochastic2024}.

While this framework was originally demonstrated on complex, high-dimensional forecasting tasks such as fluid dynamics and video prediction, we adapt it specifically for robotic LfD.
By explicitly coupling the generative source to the robot's recent state history, Temporal Policy reduces the transport distance and the high-curvature paths inherent to models initialized from independent Gaussian noise.

\section{Methodology}

\begin{figure*}[t]
   \centering
   \includegraphics[width=\textwidth]{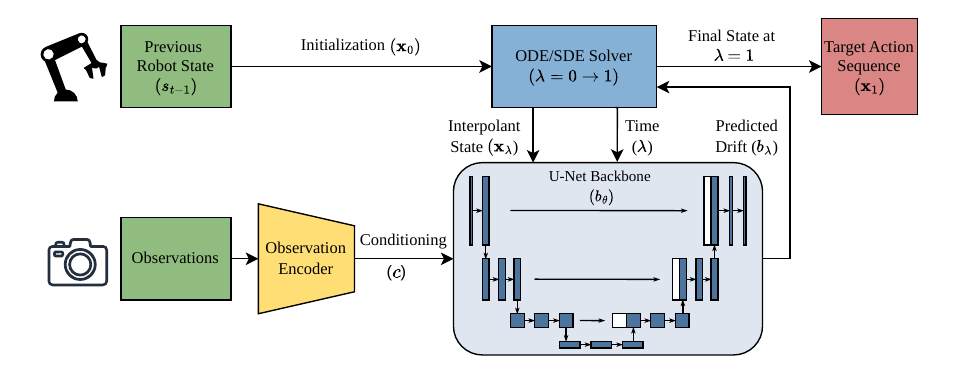}
   \caption{Overview of our architecture.
      Unlike standard generative models that initialize from independent Gaussian noise ($\mathbf{x}_0 \sim \mathcal{N}(\mathbf{0}, \mathbf{I})$, our method explicitly initializes the generative process with the robots recent state history $\mathbf{x}_0$.
      The architecture consists of: (1) An \textbf{Observation Encoder} that processes sensory observations into a conditioning vector $c$; (2) A \textbf{U-Net Backbone} that predicts the drift $b_\theta(\mathbf{x}_\lambda, \mathbf{x}_0, \lambda, c)$; and (3) an \textbf{ODE/SDE Solver} that integrates this vector field from $\lambda=0$ to $\lambda=1$ to generate the target action sequence.
   }
   \label{fig:architecture_overview}
\end{figure*}

This section details the Temporal Policy framework.
Fundamentally, continuous-time generative models learn a dynamical system that transports samples from a known source distribution $p_0(\mathbf{x}_0)$ to a complex target data distribution $p_1(\mathbf{x}_1)$.
Unlike standard diffusion and flow matching models that define $p_0$ as an uninformative Gaussian prior and map pure noise to data, we formulate action generation as a stochastic interpolant \cite{albergoStochasticInterpolantsUnifying2025} initialized directly from the robot's recent state history.
\subsection{Action Representation and Temporal Coupling}

We impose a strict structural equivalence between the state space $\mathcal{S}$ and the action space $\mathcal{A}$, defining both by the robot's configuration (e.g. joint positions or Cartesian poses).
Let $D$ denote the controlled degrees of freedom, such that $\mathcal{S} \cong \mathcal{A} \cong \mathbb{R}^D$.
The policy generates an action chunk $\mathbf{x}_1 = \{a_0, a_1, \dots, a_{H-1}\} \in \mathcal{A}^H$, over a horizon $H$.

At any given timestep $t$, we propose a natural data-dependent coupling $q_{data}(\mathbf{x}_0, \mathbf{x}_1)$ by explicitly defining the source as the robot's state history and the target as the future action sequence, shifted by $d$ discrete timesteps ($1 \le d \le H$):
\begin{itemize}
   \item \textbf{Source ($\mathbf{x}_0$):}
         State history $s_{t-H+1:t}$.
   \item \textbf{Target ($\mathbf{x}_1$):}
         Future action chunk $s_{t-H+1+d:t+d}$.
\end{itemize}
Here, the subscript notation $s_{a:b}$ denotes the sub-sequence $(s_a, s_{a+1}, \dots, s_b)$.
Sequence overlap ($d < H$) between source and target helps reduce discontinuities between action chunks.

\subsection{Static Transport Cost Analysis}
\label{sec:transport_cost}

We contrast the expected static transport cost of an independent Gaussian coupling $J_{indep}$ against our temporal coupling $J_{temporal}$.

Let the action chunk $\mathbf{x} \in \mathbb{R}^{HD}$ be a flattened sequence over horizon $H$.
Assuming the training data is standardized per degree of freedom, the expected squared $L_2$ norm evaluates to the trace of the covariance matrix, $\mathbb{E}[||\mathbf{x}||_2^2] = HD$.
For any source $\mathbf{x}_0$ and target $\mathbf{x}_1$ sharing this marginal norm, the expected transport cost expands to: $$J = \mathbb{E}[||\mathbf{x}_1 - \mathbf{x}_0||_2^2] = 2HD - 2\mathbb{E}[\mathbf{x}_1^\top \mathbf{x}_0]$$ For an independent Gaussian source, the cross-correlation term is zero ($\mathbb{E}[\mathbf{x}_1^{\top} \mathbf{x}_0] = 0$), yielding a dimension-dependent cost of $J_{indep} = 2HD$.
This necessitates learning a vector field across vast, non-physical regions of the state space, contributing to the inference latency of standard noise initialized baselines.

Conversely, our temporal coupling pairs the recent history $\mathbf{x}_0$ with the immediate future trajectory $\mathbf{x}_1$.
Because expert demonstrations tend to exhibit smooth, purposeful motions, the state vectors remain highly collinear over short temporal gaps, yielding a strong positive cross-correlation ($\mathbb{E}[\mathbf{x}_1^\top \mathbf{x}_0] > 0$).
Given this positive cross-correlation, the expected transport cost is strictly less than that of an independent coupling ($J_{temporal} < J_{indep}$).
This temporal coupling simplifies the vector field the network must learn.
\subsection{Stochastic Interpolant}
We learn the transition from $\mathbf{x}_0$ to $\mathbf{x}_1$ using the stochastic interpolants framework \cite{albergoStochasticInterpolantsUnifying2025}.
To avoid ambiguity, we explicitly distinguish between the discrete physical timestep $t$, which indexes the robot's control sequence, and the continuous interpolation time $\lambda \in [0, 1]$, which governs the generative process.
Following \cite{chenProbabilisticForecastingStochastic2024}, we define a probability path $\mathbf{x}_\lambda$ that mixes the source, target, and a standard Wiener process $\mathbf{w}_\lambda$ over the interpolation time $\lambda \in [0,1]$:
\begin{equation}
   \label{eq:stochastic_interpolant}
   \mathbf{x}_\lambda = (1-\lambda)\mathbf{x}_0 + \lambda \mathbf{x}_1 + \varepsilon(1-\lambda)\mathbf{w}_\lambda
\end{equation}
The term $\varepsilon (1-\lambda)$ represents the noise schedule where $\varepsilon \ge 0$ is a scalar hyperparameter controlling the noise scale.
We choose a linear interpolation scheme between $\mathbf{x}_0$ and $\mathbf{x}_1$ to encourage straight vector fields for fast sampling.

Because the interpolant contains a non-differentiable Wiener process, we apply It\^{o}'s calculus to derive its governing SDE:
\begin{equation}
   \dd\mathbf{x}_\lambda = \underbrace{[(\mathbf{x}_1 - \mathbf{x}_0) - \varepsilon \mathbf{w}_\lambda]}_{u_\lambda} \dd\lambda + \varepsilon(1-\lambda) \dd\mathbf{w}_\lambda
\end{equation}

A neural network $b_\theta(\mathbf{x}_\lambda, \mathbf{x}_0,  \lambda, c)$ is trained to approximate the deterministic drift target $u_\lambda$ via square-loss regression, conditioned on observations $c$:
\begin{equation}
   \mathcal{L}(\theta) = \mathbb{E}_{\lambda, (\mathbf{x}_0, \mathbf{x}_1), z} \left[ ||b_\theta - (\mathbf{x}_1 - \mathbf{x}_0 - \varepsilon\sqrt{\lambda}z)||_2^2 \right]
\end{equation}
where $z \sim \mathcal{N}(\mathbf{0}, \mathbf{I})$ is standard Gaussian noise derived from the equivalence $\mathbf{w}_\lambda \overset{d}{=} \sqrt{\lambda}z$.
The training procedure for learning the drift is detailed in Algorithm \ref{alg:training}.

\begin{algorithm}[H]
   \caption{Temporal Policy Training}
   \label{alg:training}
   \begin{algorithmic}[1]
      \STATE \textbf{Input:} Robot state-action dataset $\mathcal{D} = \{s_t, a_t\}$, Horizon $H$, Overlap $d$, Noise scale $\varepsilon$
      \REPEAT
      \STATE Sample $(\mathbf{x}_0, \mathbf{x}_1)$ where $\mathbf{x}_0 = s_{t-H+1:t}$ and $\mathbf{x}_1 = s_{t-H+1+d:t+d}$
      \STATE Sample interpolation time $\lambda \sim \mathcal{U}(0, 1)$ and noise $z \sim \mathcal{N}(\mathbf{0}, \mathbf{I})$
      \STATE Construct interpolant: $\mathbf{x}_\lambda = (1-\lambda)\mathbf{x}_0 + \lambda \mathbf{x}_1 + \varepsilon(1-\lambda)\sqrt{\lambda}z$
      \STATE Compute target drift: $u_\lambda = (\mathbf{x}_1 - \mathbf{x}_0) - \varepsilon\sqrt{\lambda}z$
      \STATE Update $\theta$ by minimizing
      \item[] $\mathcal{L}(\theta) = \|b_\theta(\mathbf{x}_\lambda, \mathbf{x}_0, \lambda, c) - u_\lambda\|^2$
      \UNTIL{converged}
   \end{algorithmic}
\end{algorithm}

\subsection{Network Architecture}
As illustrated in Figure \ref{fig:architecture_overview}, the drift vector field $b_\theta$ is parameterized using a 1D U-Net backbone \cite{ronnebergerUNetConvolutionalNetworks2015} that operates over the physical time dimension of the action chunk.
Visual observations are encoded using a ResNet-18 \cite{heDeepResidualLearning2016} and projected to 2D feature points via a Spatial Softmax layer \cite{finnDeepSpatialAutoencoders2016}.
These features are concatenated with sinusoidal embeddings of the interpolation time $\lambda$ to form the conditioning vector $c$, which modifies the U-Net feature maps via Feature-wise Linear Modulation (FiLM) layers \cite{perezFiLMVisualReasoning2018}.
While we parameterize the drift vector field $b_{\theta}$ using a standard 1D U-Net, the Temporal Policy formulation is architecture-agnostic, compatible with any backbone that supports coupled state-action dimensionality.

\subsection{Inference}
By leveraging the properties of stochastic interpolants, we can modify the noise schedule at inference time without retraining.
This is possible because the stochastic interpolant implicitly models the score function alongside the drift \cite{chenProbabilisticForecastingStochastic2024}.
Following the framework established by Chen et al., the score $\nabla_{\mathbf{x}} \log p_\lambda(\mathbf{x}|\mathbf{x}_0)$ can be recovered analytically in closed form directly from the drift $u_\lambda$ and the interpolant coefficients $\alpha_\lambda = 1-\lambda$, $\beta_\lambda = \lambda$, and $\gamma_\lambda = \varepsilon(1-\lambda)$.

The general closed-form score is given by:
\begin{equation}
   \nabla_{\mathbf{x}} \log p_\lambda(\mathbf{x}|\mathbf{x}_0) = A_\lambda [ \beta_\lambda u_\lambda - c_\lambda(\mathbf{x}, \mathbf{x}_0) ]
\end{equation}
where $A_\lambda = [\lambda\gamma_{\lambda}(\dot{\beta}_\lambda\gamma_{\lambda} - \beta_\lambda\dot{\gamma}_\lambda)]^{-1}$ and $c_\lambda(\mathbf{x}, \mathbf{x}_0) = \dot{\beta}_\lambda \mathbf{x} + (\beta_\lambda\dot{\alpha}_\lambda - \dot{\beta}_\lambda\alpha_\lambda)\mathbf{x}_0$.

For our specific interpolant defined in Equation \ref{eq:stochastic_interpolant}, these terms simplify substantially, yielding the score:
\begin{equation}
   \nabla_{\mathbf{x}} \log p_\lambda(\mathbf{x}|\mathbf{x}_0) = \frac{1}{\varepsilon^2 \lambda (1-\lambda)} \left[ \lambda u_ {\lambda} - (\mathbf{x} - \mathbf{x}_0) \right]
\end{equation}

By substituting this score into the Fokker-Planck Equation (see \cite{chenProbabilisticForecastingStochastic2024} for proof), we define an inference SDE with an arbitrary noise schedule $g_\lambda$:
\begin{equation}
   \dd\mathbf{x} = \underbrace{\left[ u_{\lambda} + \frac{g_\lambda^2 - \gamma_\lambda^2}{2} \nabla_{\mathbf{x}} \log p_\lambda(\mathbf{x}|\mathbf{x}_0) \right]}_{\tilde{b}} \dd\lambda + g_\lambda \dd\mathbf{w}
\end{equation}
where $\tilde{b}$ is the modified drift.
This enables two execution paradigms:
\begin{itemize}
   \item \textbf{Deterministic Sampling ($g_\lambda = 0$):}
         Follows a single path from source to target.
         It requires fewer steps to converge but lacks the ability to explore alternative outcomes from the same initialization.
   \item \textbf{Stochastic Sampling ($g_\lambda > 0$):}
         Injects noise to explore multimodal action trajectories.
         While this allows for diverse outputs from a single source, it typically requires more sampling steps to reach a stable solution.
\end{itemize}

\begin{algorithm}[H]
   \caption{Temporal Policy Inference}
   \label{alg:inference}
   \begin{algorithmic}[1]
      \STATE \textbf{Input:} State history $\mathbf{x}_0$, learned drift $b_\theta$, noise schedule $g_\lambda$
      \STATE \textbf{Initialize:} $\mathbf{x}_{\lambda=0} = \mathbf{x}_0$
      \FOR{$\lambda = 0$ \TO $1$ \textbf{step} $\Delta\lambda$}
      \STATE Compute score: $\nabla \log p_\lambda = \frac{1}{\varepsilon^2 \lambda (1-\lambda)} [ \lambda b_\theta - (\mathbf{x}_\lambda - \mathbf{x}_0) ]$
      \STATE Compute modified drift: $\tilde{b} = b_\theta + \frac{g_\lambda^2 - \gamma_\lambda^2}{2} \nabla \log p_\lambda$
      \IF{$g_\lambda = 0$}
      \item[] \textit{// ODE}
      \STATE Update state: $\mathbf{x}_{\lambda+\Delta\lambda} = \mathbf{x}_\lambda + \tilde{b}\Delta\lambda$
      \ELSE
      \item[] \textit{// SDE}
      \STATE Sample $z \sim \mathcal{N}(\mathbf{0}, \mathbf{I})$
      \STATE Update state: $\mathbf{x}_{\lambda+\Delta\lambda} = \mathbf{x}_\lambda + \tilde{b}\Delta\lambda + g_\lambda \sqrt{\Delta\lambda}z$
      \ENDIF
      \ENDFOR
      \STATE \textbf{Output:} Action chunk $\mathbf{x}_{\lambda=1}$
   \end{algorithmic}
\end{algorithm}

Upon recovering the final action chunk $\mathbf{x}_1$ (Algorithm \ref{alg:inference}), we employ receding horizon control, executing only the first $K$ steps ($K < H$) before querying the network for a new trajectory.
This strategy balances long-term planning with responsive execution.

\section{Experiments}
We evaluate Temporal Policy to assess its computational efficiency, geometric transport properties, and performance.
The evaluation is conducted across the Robomimic simulation benchmark and validated on a physical 7-DoF bilateral teleoperation platform.
We compare our approach against two noise initialized generative baselines: Diffusion Policy \cite{chiDiffusionPolicyVisuomotor2025} and CFM \cite{lipmanFlowMatchingGenerative2023b}.

\subsection{Simulation Experiments}
We evaluate the models on five tasks from the Robomimic benchmark \cite{mandlekarWhatMattersLearning2022}, which represent a spectrum of manipulation complexity: Lift (foundational pick-and-place), Can (pick-and-stow), Square (high-precision peg insertion), Transport (bimanual object handover), and Tool Hang (multi-stage, high-precision assembly).
To test the generative capacity of the models, we utilize both the \emph{proficient-human} (ph) datasets, which contain consistent expert demonstrations, and the \emph{multi-human} (mh) datasets (available for all tasks except Tool Hang), which introduce high variance and sub-optimal multimodal trajectories.

For all tasks, the model input includes raw RGB image observations and robot proprioception.
To ensure a rigorous comparison, we adopt the baseline training hyperparameters established by Diffusion Policy \cite{chiDiffusionPolicyVisuomotor2025}, optimizing models via AdamW with a cosine learning rate schedule and an exponential moving average of the model weights.
Models were trained with a batch size of 128 for 1500 epochs on a single NVIDIA A100 GPU.
Reported inference times were measured on a consumer NVIDIA RTX 4080 with 16GB of VRAM.

Table \ref{tab:results} summarizes the performance.
All models achieve comparably high success rates overall, though Temporal Policy shows slight task-specific variations, outperforming baselines on Can mh, Square ph, and Transport mh, but underperforming on Square mh and Tool Hang ph.
\begin{table}[t]
   \centering
   \caption{Performance on simulation tasks.
      We report \emph{Max/Average} success rates.
      \emph{Max} is the maximum rate over all training, and \emph{Average} is the mean of the last 10 evaluation checkpoints.
      Each checkpoint includes 50 task evaluations and all values are averaged across 3 seeds.
      'Ph' refers to \emph{proficient-human}, and 'mh' refers to \emph{multi-human}.
      Bold indicates best results.
   }
   \label{tab:results}
   \begin{tabular}{ll | ccc}
      \toprule
      \multicolumn{2}{l|}{Task}           & Diffusion Policy & CFM                         & Temporal Policy                                           \\
      \midrule
      \multirow{2}{*}{Lift}               & ph               & \textbf{1.00}/\textbf{1.00} & \textbf{1.00}/\textbf{1.00} & \textbf{1.00}/\textbf{1.00} \\
                                          & mh               & \textbf{1.00}/\textbf{1.00} & \textbf{1.00}/0.99          & \textbf{1.00}/0.99          \\
      \multirow{2}{*}{Can}                & ph               & \textbf{1.00}/0.97          & \textbf{1.00}/\textbf{0.98} & \textbf{1.00}/\textbf{0.98} \\
                                          & mh               & 0.99/0.96                   & 0.99/0.95                   & \textbf{1.00}/\textbf{0.98} \\
      \multirow{2}{*}{Square}             & ph               & 0.98/0.91                   & 0.97/0.91                   & \textbf{0.99}/\textbf{0.96} \\
                                          & mh               & \textbf{0.93}/\textbf{0.87} & 0.85/0.78                   & 0.91/0.83                   \\
      \multirow{2}{*}{Transport}          & ph               & 0.97/0.89                   & \textbf{0.98}/\textbf{0.91} & 0.97/\textbf{0.91}          \\
                                          & mh               & 0.82/0.73                   & 0.67/0.54                   & \textbf{0.85}/\textbf{0.76} \\
      \multirow{1}{*}{Tool Hang}          & ph               & \textbf{0.87}/\textbf{0.71} & 0.79/0.70                   & 0.81/0.69                   \\
      \midrule
      \multicolumn{2}{l|}{NFE}            & 100              & \textbf{10}                 & \textbf{10}                                               \\
      \multicolumn{2}{l|}{Inference (ms)} & 615.6            & 63.5                        & \textbf{19.1}                                             \\
      \bottomrule
   \end{tabular}
\end{table}
Like CFM, our method utilizes only 10 function evaluations (NFE) to generate action sequences.
While the baseline Diffusion Policy and CFM models in our evaluation employ 255M parameters, we implement Temporal Policy using a compact 17M parameter architecture.
We found that the full 255M parameter model led to overfitting for our Temporal Policy.
However, a more compact network may also be sufficient for other baselines.
We trained a scaled-down CFM baseline restricted to the same 17M parameter architecture on the transport task and did not observe degraded performance.
For Temporal Policy, the more compact network and low NFE together result in a reduced inference time of 19.1 ms.

\begin{figure}[t]
   \centering
   \includegraphics[width=\linewidth]{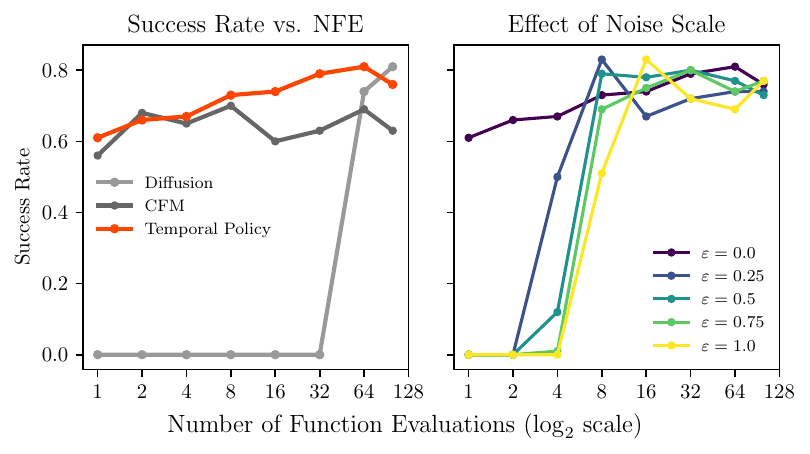}
   \caption{Comparison of model performance on the Tool Hang ph task. (Left) Success rate vs.
      NFE for Diffusion, CFM, and Temporal Policy.
      (Right) An ablation study showing the effect of the noise scale on success rate.}
   \label{fig:performance_results}
\end{figure}

This low NFE is enabled by the geometry of the generative flow.
By initializing from the robot's history, Temporal Policy reduces the transport distance to the target action sequence.
As illustrated in Figure \ref{fig:performance_results} (Left), standard diffusion fails to produce valid actions below 32 NFEs, requiring at least 64 steps to reach peak performance.
Conversely, Temporal Policy achieves relatively high success rates with only a single sampling step.
This fast convergence empirically validates that the data-dependent coupling simplifies the target vector field, allowing low-order ODE solvers to take large step sizes without accumulating fatal discretization errors.

\subsection{Transport Cost}
To quantify the geometric efficiency of the generative process, we compute the dynamic transport cost $\mathcal{T}$, defined as the path integral of the generated sample's evolution in the configuration space.
For a discrete integration trajectory $\tau = \{x_0, \dots, x_K\}$ where $x_k \in SE(3)$ represents the sample state at step $k$, the transport cost is:

\begin{equation}
   J(\tau) = \sum_{k=0}^{K-1} \left( \| \mathbf{p}_{k+1} - \mathbf{p}_k \|_2 + r \cdot \theta(R_k, R_{k+1}) \right)
\end{equation}

Here, $\mathbf{p}$ denotes translation, $R$ denotes the rotation matrix, and $\theta(R_a, R_b)$ represents the geodesic distance between rotations.
The characteristic length $r$ is set to $0.1$ m to balance units.

\begin{table}[h]
   \centering
   \caption{Transport metrics on the Square ph task (averaged over 50 episodes).
   }
   \label{tab:transport}

   \begin{tabular} {c c c}
      \toprule
      Model            & Transport     & Straightness  \\
      \midrule
      Diffusion Policy & 189.99        & 20.61         \\
      CFM              & 10.00         & 1.08          \\
      Temporal Policy  & \textbf{1.21} & \textbf{1.02} \\
      \bottomrule
   \end{tabular}

\end{table}

As detailed in Table \ref{tab:transport}, our Temporal Policy exhibits markedly smaller transport costs compared to baselines.
We attribute this reduction to two factors: (1) the proximity of initialization, where starting from the robot's past history places the source distribution significantly closer to the target action than standard Gaussian noise; and (2) the linearity of the flow, induced by the linear interpolant objective.

To disentangle these effects, we define the Straightness Ratio to quantify the deviation of the learned flow from a straight-line path:
\begin{equation}
   \frac{J(\tau)}{\| \mathbf{p}_{K} - \mathbf{p}_0 \|_2 + r \cdot \theta(R_0, R_K)}
\end{equation}

A straightness ratio $ \approx 1$ implies the model moves particles along straight lines in the configuration manifold.
Temporal Policy achieves a ratio of $1.02$, significantly outperforming Diffusion Policy ($20.61$), with only slight improvements over CFM ($1.08$).
This confirms that utilizing the robot's history as the flow source not only shortens the generation distance but also induces straight paths, directly enabling the efficient, low-NFE sampling observed in Figure \ref{fig:performance_results} (Left).

Comparison with CFM, which exhibits mostly straight paths, reveals our method's lower transport cost mostly stems from starting closer to the target manifold, rather than from straightening the flow.

\subsection{Ablation Study: SDE Noise Scheduling}
To determine the optimal inference strategy, we evaluate the practical trade-off between computational latency (number of integration steps) and task performance.
We compare the deterministic Probability Flow ODE ($\varepsilon=0.0$) against an SDE with varying noise scales ($\varepsilon > 0$) on the Tool Hang ph task.

As shown in Figure \ref{fig:performance_results} (Right), the dominant effect of increasing the diffusion coefficient is a rightward shift in success rates, demonstrating a penalty on required sampling steps.
The deterministic ODE ($\varepsilon=0.0$) achieves a relatively high success rates in as few as 2 to 4 integration steps.
Conversely, introducing stochasticity severely degrades low-step performance and fully stochastic sampling ($\varepsilon=1.0$) requires at least 8 to 16 steps to reach parity.

This rightward shift reflects the weaker convergence order of the Euler-Maruyama integration required for SDEs compared to the deterministic ODE solver.
Since stochastic sampling does not yield higher success rates to justify its computation overhead, the deterministic ODE is the most efficient choice.

\subsection{Data Efficiency}
\begin{figure}[h]
   \centering
   \includegraphics[width=\linewidth]{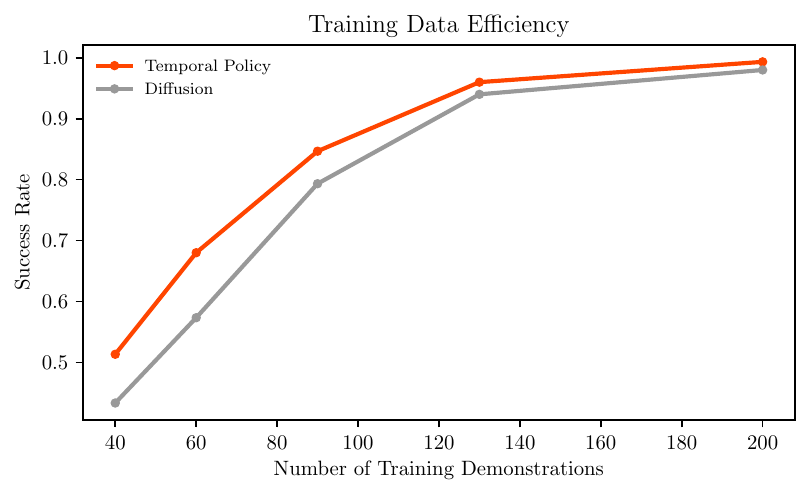}
   \caption{Data efficiency on the Square ph task.}
   \label{fig:data_efficiency}
\end{figure}

Figure \ref{fig:data_efficiency} illustrates success rates as a function of dataset size.
Temporal Policy consistently outperforms Diffusion Policy across all data regimes.
We hypothesize that the Temporal Policy formulation simplifies the learning problem by reducing the discrepancy between the source and target distributions, allowing the model to generalize effectively with fewer demonstrations.
This efficiency has important practical implications, as the high cost of data collection remains a significant barrier to the adoption of LfD methods.

\subsection{Real-World Experiments}
We validated the framework on a teleoperation platform comprising two Barrett WAM manipulators configured in a bilateral leader-follower architecture.
The leader device is a 4-DoF Barrett WAM equipped with a custom-built 3-DoF wrist exoskeleton, providing a full 7-DoF interface.
This hybrid configuration allows the human operator to control the full pose (position and orientation) of the end-effector.
The follower is a standard 7-DoF Barrett WAM manipulator mounted with a BarrettHand (BH8-262) end-effector for grasping.
The system operates on a 500 Hz control loop, where the first 4 DoF utilize bilateral position-position control and the 3-DoF wrist utilizes bilateral orientation control, providing full 7-DoF force feedback to the operator.
Visual observations are provided by two FLIR Blackfly S USB3 cameras streaming at 30 FPS, configured with one fixed third-person view and one wrist-mounted view.

\begin{figure}[t]
   \centering
   \begin{subfigure}{0.48\linewidth}
      \includegraphics[width=\linewidth]{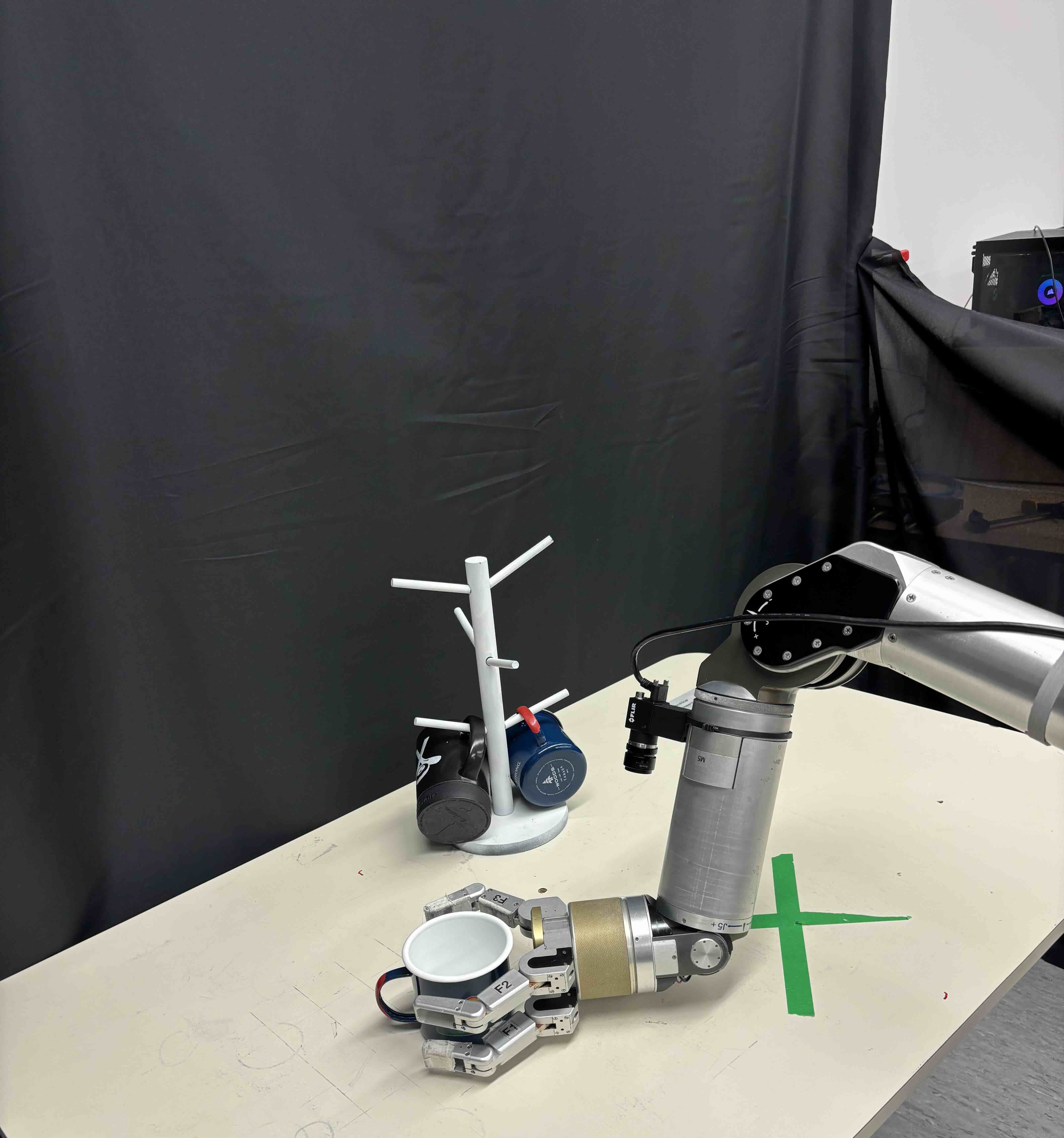}
      \caption{Start State}
      \label{fig:task_start}
   \end{subfigure}
   \hfill
   \begin{subfigure}{0.48\linewidth}
      \includegraphics[width=\linewidth]{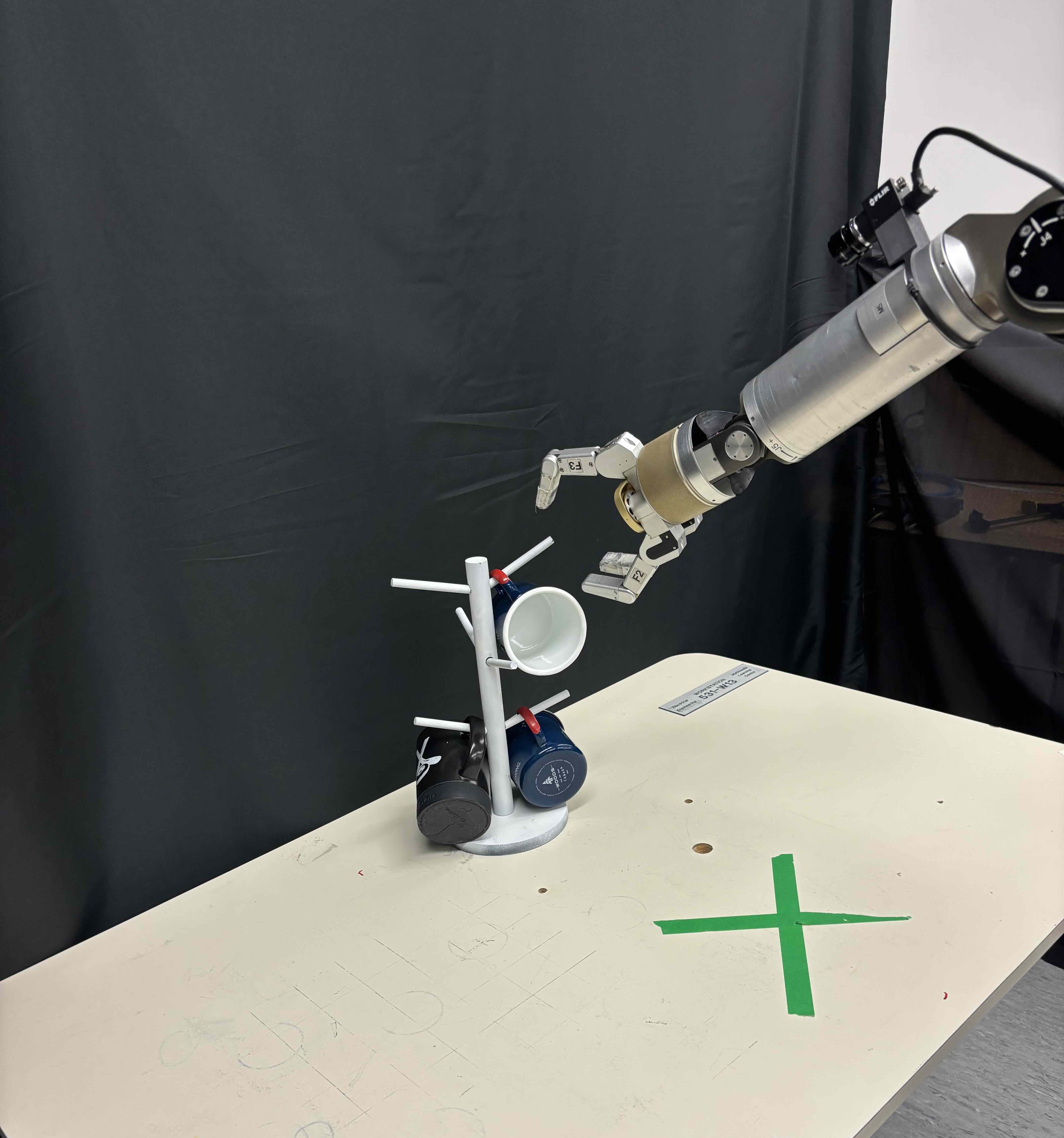}
      \caption{End State}
      \label{fig:task_end}
   \end{subfigure}
   \caption{Overview of the mug-hang task. (a) The initial configuration. (b) The successful terminal state.}
   \label{fig:mug_task_overview}
\end{figure}

We evaluated the policy on a household-inspired manipulation task, hanging a mug on a tree (Figure \ref{fig:mug_task_overview}).
The robot must first grasp an unconstrained mug such that the handle faces away from the gripper palm, orient the mug, and align the handle to hang it from a fixed tree peg.
This task specifically evaluates the policy's ability to handle unsupported free-space alignment, where slight depth or angular deviations result in missed connections.

To train the policy, we collected 150 expert demonstrations of the hanging task using the bilateral teleoperation interface.
The state-action temporal coupling consisted of follower joints positions and gripper velocities, utilizing an overlap of 2 between source and target pairs ($\mathbf{x}_0, \mathbf{x}_1)$.
To provide necessary context, the model was additionally conditioned on the gripper joint positions and the RGB images with an observation window of 2.
During autonomous physical deployment, the policy operates at a control frequency of 10 Hz.
At each inference step, the model predicts a future action trajectory of 16 steps while only executing 8.
For fast inference, we used deterministic sampling with 10 Euler steps.

\subsection{Real-World Results}

To validate real-world applicability, we deployed Temporal Policy on the physical hardware for 20 evaluation trials of the mug-hanging task.
Across the 20 trials with randomized initial mug configurations, Temporal Policy achieved a 50\% (10/20) overall success rate.
The policy demonstrated high robustness in the initial acquisition phase, successfully grasping the mug in 95\% (19/20) episodes.
The primary failure mode occurred during the alignment phase, where the end-effector systematically undershot the depth of the tree peg.
We hypothesize this is caused by the grasped mug occluding the target peg from the wrist camera.

\section{CONCLUSIONS}

Standard noise initialized generative models for LfD incur high transport costs and inference latency due to their reliance on uninformative Gaussian priors.
This paper introduced Temporal Policy, an alternative generative framework that reframes action chunking as a Point-to-Distribution transport problem initialized directly from the robot's state history.

By coupling past states to future actions, Temporal Policy demonstrates reduced transport costs and straight vector fields.
This geometric efficiency allows the network to match state-of-the-art success rates on visuomotor manipulation tasks while achieving a 19.1~ms inference latency on consumer grade hardware.
Our results also suggest improved data efficiency, enabling the model to achieve higher success rates with fewer demonstrations than noise-based baselines.
We demonstrated the feasibility of this approach in a real-world deployment on a 7-DoF manipulator, achieving a 95\% grasp success rate and a 50\% overall success rate on a mug-hanging task.
Furthermore, the interpolant formulation supports both deterministic (ODE) and stochastic (SDE) sampling within a single model.

A primary limitation of this point-to-distribution formulation is its sensitivity to initialization quality; significant noise at the source state can bias the generated trajectory.
Future work should explore whether history initialization primarily disambiguates apparent multimodality, and whether it preserves the full range of valid behaviors observed in the demonstrations.
To further minimize transport cost, subsequent research will explore using the robot's current velocity to project its starting state forward, initializing the flow closer to the target manifold.
Finally, we plan to investigate alternative state-action representations, such as force/torque control or keypoint generation for visual servoing.

Because Temporal Policy modifies the data-dependent coupling, our framework is orthogonal to many concurrent advancements in generative modeling.
Future iterations can leverage consistency distillation \cite{songConsistencyModels2023} or higher-order numerical solvers to further reduce the required number of function evaluations.

\addtolength{\textheight}{-0.8cm}

\section*{ACKNOWLEDGMENT}
This research was supported by \censor{Natural Sciences and} \censor{Engineering Research Council of Canada [23-04887]}, \censor{Department of National Defense, [RRU08]} and the \censor{Canada Foundation for Innovation}.

\bibliographystyle{IEEEtran}

\bibliography{refs.bib}

\end{document}